\documentclass[sigconf,nonacm]{acmart}

\AtBeginDocument{%
}

\usepackage{multirow}
\usepackage{tikz}

\newcommand{\circnum}[1]{%
  \tikz[baseline=(char.base)]{%
    \node[draw,circle,inner sep=1pt] (char)
      {\footnotesize\bfseries #1};%
  }%
}

\begin{document}

\title{Don't Overthink, Don't Underthink: \\Toward Adaptive Reasoning in Agentic AI}

\author{Md Jueal Mia and M. Hadi Amini}
\email{\{mmia001, moamini\}@fiu.edu}
\affiliation{%
  \institution{Knight Foundation School of Computing and Information Sciences,\\Security, Optimization, and Learning for InterDependent Networks Laboratory (solid lab)\\Florida International University}
  \city{Miami}
  \state{Florida}
  \country{USA}
}

\renewcommand{\shortauthors}{Mia and Amini}

\begin{abstract}
Recent advances in Large Language Models (LLMs) have shown that increased inference-time reasoning can improve performance on complex tasks. However, many existing approaches rely on fixed or preallocated reasoning controls, such as fixed token budgets, pre-execution difficulty estimates, or activation-space interventions, and are often evaluated on standalone reasoning benchmarks rather than full agentic workflows. These assumptions may not hold in agentic AI systems, where reasoning requirements evolve dynamically through planning, tool use, memory retrieval, and agent-to-agent interactions. Consequently, reasoning can become either excessive or insufficient, resulting in unnecessary computation, increased latency, planning drift, excessive tool use, or incomplete solutions. We argue that a major challenge for next-generation agentic AI is not merely how much reasoning a language model should perform, but how it should allocate reasoning according to evolving task demands. We characterize over-reasoning and under-reasoning as recurring failure modes of misallocated reasoning and evaluate them on MATH-500 and the GAIA public validation benchmark. Using tool-decision latency, token consumption, token-limit exhaustion, and answer correctness, our results suggest that cases classified as over-reasoning are associated with higher computational cost without proportional accuracy gains, whereas cases classified as under-reasoning are consistently associated with incorrect or incomplete solutions. These findings motivate future research on adaptive reasoning mechanisms for agentic AI.

\end{abstract}

\keywords{agentic AI, adaptive reasoning, over-reasoning, under-reasoning, reasoning efficiency}

\maketitle

\section{Introduction}

Recent advancements in LLMs have significantly enhanced their reasoning capabilities, enabling them to solve complex problems through step-by-step deliberation. This progress has led to the emergence of a class of models known as Large Reasoning Models (LRMs), which are typically developed through supervised fine-tuning (SFT) and reinforcement learning from human feedback (RLHF). By generating chain-of-thought (CoT) reasoning traces, LRMs can perform sophisticated tasks involving logical deduction, mathematical reasoning, planning, and decision-making \cite{sui2025stop}. As a result, these models have demonstrated strong performance across a wide range of challenging benchmarks. These advances build upon a series of foundational developments, including Chain-of-Thought prompting for explicit reasoning \cite{wei2022chain}, reasoning-and-acting frameworks such as ReAct \cite{yao2022react} and the emergence of dedicated reasoning models including OpenAI o1 \cite{openai2024o1} and Phi-4-reasoning \cite{abdin2025phi}.

Despite these improvements, explicit reasoning introduces substantial computational overhead during inference by increasing latency and token consumption, leading to higher inference costs. Recent surveys identify efficient inference as a key challenge for next-generation language models, emphasizing the need for mechanisms that better allocate computational resources during inference \cite{amini2025distributed}. Although extensive reasoning is often beneficial for complex tasks, it is frequently unnecessary for simpler queries, and existing LRMs often allocate reasoning inefficiently by over-computing on easy problems while under-reasoning on more difficult ones, creating a trade-off between reasoning quality and computational efficiency \cite{aggarwal2025optimalthinkingbench}. This challenge becomes even more pronounced in agentic AI systems, where one or more LLMs coordinate planning, tool use, memory retrieval, environment interaction, and multi-step decision making \cite{sapkota2025ai}. In such systems, specialized components (e.g., planners and response generators) may independently invoke LLMs and perform their own reasoning processes, causing unnecessary reasoning at individual components to accumulate across the pipeline and significantly increase overall latency, token usage, and computational cost.


\begin{figure*}[t]
    \centering
    \includegraphics[width=0.72\textwidth]{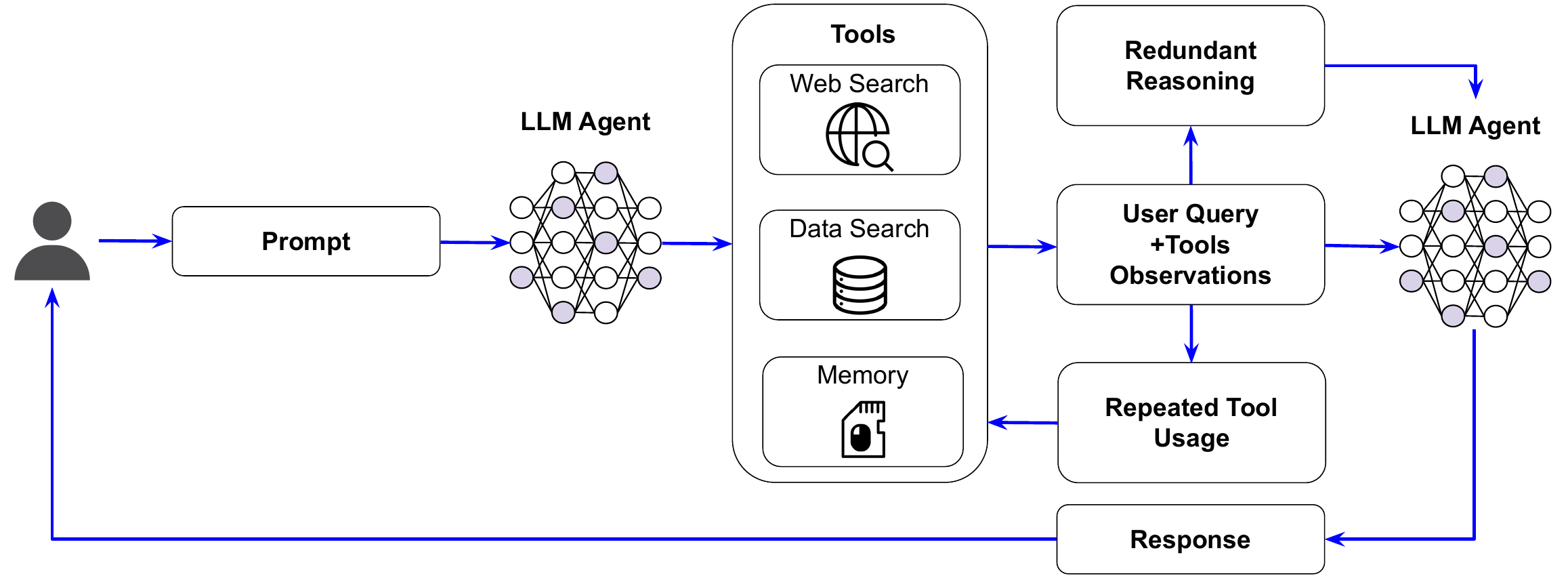}
    \caption{Overview of over-reasoning and under-reasoning in agentic AI. The agent loop is annotated with four failure modes: redundant reasoning, weak reasoning, repeated tool usage, and no tool usage (skipped tool calls).}
    \label{fig:agentic_overview}
\end{figure*}

As inference-time computation becomes a important factor in the deployment of modern AI systems, understanding how to dynamically adapt reasoning effort to the complexity of a given task has emerged as an important research challenge. Developing mechanisms that allocate reasoning resources more efficiently can reduce unnecessary computational costs while maintaining, or even improving, overall task performance. This challenge is particularly important for agentic AI systems, where reasoning efficiency directly impacts
scalability, responsiveness, and real-world usability. Figure~\ref{fig:agentic_overview} illustrates the major challenges that arise from mis-allocated reasoning that motivate this work. The main contributions of our work are listed below.

\begin{itemize}

\item We investigate the impact of over-reasoning and under-reasoning in agentic AI, showing how reasoning effort that is either excessive or insufficient relative to task requirements is associated with higher inference cost or degraded task performance.

\item We present a quantitative analysis of reasoning allocation in a LangGraph-based agentic AI framework using representative reasoning models on MATH-500 and GAIA. Our results reveal that current reasoning models frequently exhibit over-reasoning or under-reasoning, resulting in inefficient reasoning allocation for agentic AI.

\item Based on our findings, we identify key research challenges and outline candidate evaluation dimensions and future directions for adaptive reasoning mechanisms that dynamically determine when to reason, when to act, and when to stop.

\end{itemize}

\section{Related Works}

Recent studies have explored the growing role of reasoning in agentic AI systems. Wu et al. \cite{wu2025agentic} demonstrated that integrating search, coding, and memory agents can enhance complex reasoning capabilities . As agentic reasoning becomes increasingly sophisticated, researchers have begun examining its efficiency implications. Kim et al. \cite{kim2026cost} showed that deeper reasoning and test-time scaling introduce substantial computational and infrastructure costs. Similarly, Tran et al. \cite{tran2026single} found that increasing reasoning and coordination does not always improve performance, with single-agent systems often outperforming multi-agent architectures under equal compute budgets. Wang et al. \cite{wang2025efficient} showed that increasing reasoning budgets, adding more planning steps, and employing complex memory mechanisms often result in only marginal performance improvements while significantly increasing computational cost, while Becker et al. \cite{becker2026stay} revealed that prolonged agent interactions can cause discussions to drift away from task objectives, leading to performance degradation. Collectively, these findings suggest that increasing reasoning effort, agent interactions, and test-time computation does not necessarily translate into better agent performance and may instead introduce inefficiencies such as overthinking, planning drift, and coordination overhead.

Despite this progress, many existing approaches still rely on fixed reasoning budgets or evaluate adaptive reasoning outside full agentic workflows. These assumptions may not hold in agentic AI, where reasoning requirements evolve dynamically through interactions with tools, memory, and external environments. As a result, agents must determine when to continue reasoning, invoke tools, or act, while additional deliberation may amplify noisy observations rather than improve decisions. Moreover, prior work has shown that prolonged reasoning and replanning can lead to planning drift, where agents gradually deviate from the user's objective, although planning drift is not directly measured in this study. These observations raise a key question: when does additional reasoning improve agent performance, and when does it become counterproductive? Answering this question is essential for developing agentic AI systems that allocate reasoning efficiently while maintaining alignment with task objectives.

\section{Preliminaries}

\subsection{Agentic AI Systems}

Agentic AI represents a shift from traditional AI systems toward autonomous, goal-oriented systems capable of operating in dynamic environments with minimal human intervention \cite{acharya2025agentic}. Unlike conventional AI, which is typically designed for predefined tasks, agentic AI can adapt to changing conditions, make decisions based on contextual information, and pursue complex objectives over extended periods \cite{hosseini2025role}. Recent surveys further characterize agentic AI as an evolution beyond single-agent architectures, emphasizing multi-agent collaboration, dynamic task decomposition, persistent memory, and coordinated autonomy to accomplish complex goals across diverse application domains \cite{sapkota2025ai}. These capabilities enable agentic systems to move beyond simple automation toward autonomous problem solving, adaptive decision-making, and collaborative task execution in real-world environments.

\subsection{Under-Reasoning, Adequate Reasoning, and Over-Reasoning}

Recent advances in reasoning models show that increasing reasoning effort does not always improve performance. Models may allocate reasoning inefficiently, producing unnecessarily long reasoning traces for simple problems while failing to devote sufficient reasoning to more challenging ones, and answer correctness often exhibits a non-linear relationship with reasoning length \cite{su2025between}. In agentic settings, stronger reasoning improves planning-oriented tasks but incurs substantially higher computational costs and reasoning overhead \cite{zhou2025exploring}. Moreover, excessive test-time reasoning yields diminishing returns and may even cause models to abandon previously correct answers, highlighting the risks of overthinking during inference \cite{zhou2026more}. These findings suggest that agentic systems should adapt reasoning effort to task complexity rather than relying on uniformly long reasoning processes.

Conceptually, we distinguish three reasoning regimes. \emph{Under-reasoning} occurs when an agent allocates insufficient reasoning, evidence gathering, or verification for the task, such as omitting intermediate reasoning steps, failing to gather sufficient evidence, or terminating prematurely. \emph{Over-reasoning} occurs when an agent continues reasoning beyond what is necessary, exhibiting redundant deliberation, repeated verification, or unnecessary reasoning that increases computational cost. Between these extremes lies \emph{adequate reasoning}, where reasoning effort is sufficient for the task without exhibiting either under-reasoning or over-reasoning.

\section{Preliminary Study: Evidence of Misallocated Reasoning}

\subsection{Experimental Setup}
Experiments were conducted on a server with two NVIDIA RTX A6000 GPUs (48,GB VRAM each). Our LangGraph-based agentic AI framework was served through vLLM-compatible APIs using Qwen3.5-4B as the router,\footnote{Hugging Face identifier: \nolinkurl{Qwen/Qwen3.5-4B}.} while the final-response models were Qwen3.5-4B, Llama-3.1-8B-Instruct,\footnote{Hugging Face identifier: \nolinkurl{meta-llama/Llama-3.1-8B-Instruct}.} and Phi-4-reasoning.\footnote{Hugging Face identifier: \nolinkurl{microsoft/Phi-4-reasoning}.} The Qwen3.5-4B configuration used the same model for both router and final response. Each model was evaluated on all 500 MATH-500 test problems~\cite{math500} and 165 GAIA validation tasks~\cite{mialon2024gaia} using deterministic decoding (temperature $=0.0$) with a maximum generation length of 4,096 tokens. We recorded tool usage, latency, token consumption, token-limit hits, and complete interaction traces. For reasoning models, reasoning tokens were computed from tokens enclosed by \texttt{<think>} and \texttt{</think>}; since Llama-3.1-8B-Instruct does not expose reasoning traces, this metric is reported as \texttt{N/A}. Finally, GPT-4.1 independently classified each trajectory as \emph{over-reasoning}, \emph{under-reasoning}, or \emph{adequate reasoning}, while final answers were labeled as \emph{correct}, \emph{incorrect}, or \emph{incomplete}, with reasoning labels assigned independently of answer correctness.

\section{Findings}
\begin{table*}[t]
\centering
\caption{Average runtime and reasoning statistics in agentic AI across MATH-500 and GAIA. Tool agent model: Qwen3.5-4B.}
\label{tab:runtime_reasoning_summary}

\resizebox{\textwidth}{!}{
\begin{tabular}{llcccccccccc}
\hline
\textbf{Dataset} &
\textbf{Final Model} &
\textbf{Avg. Total Time (s)} &
\textbf{Avg. Tool Decision Time (s)} &
\textbf{Avg. Final Answer Time (s)} &
\textbf{Avg. Tool Calls/Sample} &
\textbf{Avg. Tool Input Tokens} &
\textbf{Avg. Tool Output Tokens} &
\textbf{Avg. Final Output Tokens} &
\textbf{Avg. Final Reasoning Tokens} &
\textbf{Token Limit Hits} \\
\hline

\multirow{3}{*}{MATH-500}
& Qwen3.5-4B             & 105.76 & 34.95 & 70.21  & 00.39 & 583.30 & 1224.89 & 2461.67 & 1732.39 & 190 \\
& Llama-3.1-8B-Instruct  & 57.76  & 35.75 & 21.42  & 00.39 & 583.39 & 1225.09 & 904.48  & N/A     & 71 \\
& Phi-4-reasoning        & 216.46 & 35.08 & 180.80 & 00.39 & 583.30 & 1224.89 & 4046.59 & 2220.12 & 482 \\

\hline

\multirow{3}{*}{GAIA}
& Qwen3.5-4B             & 79.64  & 10.92 & 67.49  & 00.93 & 582.24 & 379.53 & 2353.90 & 1219.76 & 75 \\
& Llama-3.1-8B-Instruct  & 22.77  & 10.94 & 10.71  & 00.93 & 582.24 & 379.53 & 452.23  & N/A     & 10 \\
& Phi-4-reasoning        & 195.00 & 10.74 & 182.94 & 00.93 & 582.24 & 379.53 & 4091.12 & 2897.95 & 164 \\

\hline
\end{tabular}
}
\end{table*}

\begin{table*}[t]
\centering
\caption{Reasoning efficiency and answer quality in agentic AI across MATH-500 and GAIA. Tool agent model: Qwen3.5-4B. Judge model: GPT-4.1. Hit Accuracy denotes accuracy on samples where the final model reached the maximum output token limit, while No-Hit Accuracy denotes accuracy on samples that did not reach the maximum output token limit.}
\label{tab:reasoning_efficiency}

\resizebox{\textwidth}{!}{
\begin{tabular}{llcccccccccccc}
\hline
\textbf{Dataset} &
\textbf{Final Model} &
\textbf{Accuracy(\%)} &
\textbf{Answer Correct} &
\textbf{Answer Incorrect} &
\textbf{Incomplete} &
\textbf{Token Limit Hits} &
\textbf{Hit Accuracy(\%)} &
\textbf{No-Hit Accuracy(\%)} &
\textbf{Over-Reasoning} &
\textbf{Under-Reasoning} &
\textbf{Adequate Reasoning} &
\textbf{Over-Reasoning Rate(\%)} &
\textbf{Under-Reasoning Rate(\%)} \\
\hline

\multirow{3}{*}{MATH-500}
& Qwen3.5-4B             & 87.60 & 438 & 14  & 48  & 190 & 73.68 & 96.13 & 340 & 40  & 120 & 68.00 & 8.00 \\
& Llama-3.1-8B-Instruct  & 49.80 & 249 & 187 & 64  & 71  & 9.86  & 56.41 & 99  & 110 & 291 & 19.80 & 22.00 \\
& Phi-4-reasoning        & 92.40 & 462 & 5   & 33  & 482 & 92.12 & 100.00 & 447 & 24  & 29  & 89.40 & 4.80 \\

\hline

\multirow{3}{*}{GAIA}
& Qwen3.5-4B             & 12.12 & 20 & 79  & 66  & 75  & 13.33 & 11.11 & 63 & 61  & 41 & 38.18 & 36.97 \\
& Llama-3.1-8B-Instruct  & 5.45  & 9  & 100 & 56  & 10  & 0.00  & 5.81  & 17 & 118 & 30 & 10.30 & 71.52 \\
& Phi-4-reasoning        & 11.52 & 19 & 16  & 130 & 164 & 11.59 & 0.00 & 58 & 103 & 4 & 35.15 & 62.42 \\

\hline
\end{tabular}
}
\end{table*}

Analysis of Tables~\ref{tab:runtime_reasoning_summary} and \ref{tab:reasoning_efficiency} suggests that both over-reasoning and under-reasoning emerge as recurring failure modes in the evaluated LangGraph-based agent across reasoning-intensive (MATH-500) and real-world multi-step (GAIA) tasks. Rather than consistently allocating reasoning effort according to task requirements, the evaluated agent frequently allocates either excessive or insufficient reasoning, resulting in increased computational cost or degraded task performance.

The first major issue is \textbf{over-reasoning}. On MATH-500, Phi-4-reasoning exhibits the highest over-reasoning rate (89.40\%), followed by Qwen3.5-4B (68.00\%) and Llama-3.1-8B-Instruct (19.80\%). It also generates the most reasoning tokens (2220.12), the longest outputs (4046.59 tokens), and the highest final inference time (180.80\,s), compared with Qwen3.5-4B (1732.39 reasoning tokens, 2461.67 output tokens, 70.21\,s) and Llama-3.1-8B-Instruct (904.48 output tokens, 21.42\,s). Although Phi-4-reasoning achieves the highest accuracy (92.40\%), it reaches the maximum token limit in 482 of 500 samples, compared with 190 and 71 for Qwen3.5-4B and Llama-3.1-8B-Instruct, respectively. A similar trend is observed on GAIA, where Phi-4-reasoning again produces the longest reasoning traces (2897.95 reasoning tokens, 4091.12 output tokens, 182.94\,s). Since Llama-3.1-8B-Instruct does not expose explicit reasoning traces, its reasoning-token count is reported as \texttt{N/A}. Overall, increasing explicit reasoning substantially increases inference cost.

The second major issue is that \textbf{over-reasoning does not necessarily translate into proportional performance gains}. On MATH-500, Phi-4-reasoning improves accuracy over Qwen3.5-4B (92.40\% vs.\ 87.60\%) but nearly triples final inference time (180.80\,s vs.\ 70.21\,s), increases output length (4046.59 vs.\ 2461.67 tokens), and more than doubles token-limit hits (482 vs.\ 190). On GAIA, despite producing much longer reasoning traces and requiring almost three times longer inference than Qwen3.5-4B (182.94\,s vs.\ 67.49\,s), Phi-4-reasoning achieves slightly lower accuracy (11.52\% vs.\ 12.12\%). These observations indicate that substantially increasing reasoning effort does not necessarily yield proportional performance improvements.

The third major issue is \textbf{under-reasoning}. On MATH-500, Llama-3.1-8B-Instruct exhibits the highest under-reasoning rate (22.00\%), followed by Qwen3.5-4B (8.00\%) and Phi-4-reasoning (4.80\%), resulting in 64, 48, and 33 incomplete responses, respectively. On GAIA, under-reasoning increases to 71.52\% for Llama-3.1-8B-Instruct, 62.42\% for Phi-4-reasoning, and 36.97\% for Qwen3.5-4B, corresponding to 56, 130, and 66 incomplete responses. Although reasoning categories are assigned independently of correctness, all under-reasoning cases were associated with incorrect or incomplete answers, suggesting that reasoning often terminates before collecting sufficient evidence or adequately verifying conclusions.

The fourth major issue is \textbf{token budget exhaustion}. On MATH-500, Phi-4-reasoning reaches the maximum output token limit in 482 of 500 samples (96.40\%), substantially exceeding Qwen3.5-4B (190; 38.00\%) and Llama-3.1-8B-Instruct (71; 14.20\%). Despite this, its hit accuracy remains high (92.12\%), indicating that many responses are already correct before reaching the limit. On GAIA, Phi-4-reasoning again records the most token-limit hits (164 of 165), compared with 75 and 10 for Qwen3.5-4B and Llama-3.1-8B-Instruct, while hit accuracy remains low for both reasoning models (11.59\% and 13.33\%), suggesting that exhausting the token budget does not substantially improve performance on complex tasks.

The fifth major issue is \textbf{task-dependent reasoning and tool-use allocation}. On MATH-500, the agent averages 0.39 tool calls per sample, consuming approximately 583 tool-input tokens and 1225 tool-output tokens despite external information rarely being required. In contrast, GAIA averages 0.93 tool calls per sample because external information and multi-step interactions are frequently necessary. Nevertheless, despite greater tool use, the highest GAIA accuracy remains only 12.12\%, suggesting that increasing reasoning length and tool interactions alone is insufficient to substantially improve task performance. Overall, the evaluated LangGraph-based agent does not consistently allocate reasoning effort or tool use efficiently across different task types.
\section{Discussion and Future Directions}

Our findings show that reasoning efficiency remains a key challenge for both current LRMs and agentic AI systems. On simpler reasoning tasks, reasoning models can improve accuracy but often incur substantially higher latency, token consumption, and token-budget exhaustion. On more complex agentic tasks, additional reasoning greatly increases computational cost without corresponding accuracy gains, while insufficient reasoning often results in incorrect or incomplete solutions. These challenges become more pronounced in agentic AI, where reasoning is intertwined with planning, tool use, memory retrieval, and multi-step decision making. Together, these observations motivate four key research challenges:
\textbf{\circnum{1} Unknown Reasoning Sufficiency}, where agents cannot determine when enough reasoning has been performed and should stop, leading to unnecessary computation; \textbf{\circnum{2} Over-Reasoning and Planning Drift}, where excessive reasoning increases computational cost and may introduce redundant deliberation, planning drift, or incorrect intermediate reasoning without improving answer quality; \textbf{\circnum{3} Under-Reasoning and Premature Decisions}, where agents fail to recognize when additional reasoning, evidence gathering, or verification is required before producing a final response; and \textbf{\circnum{4} Dynamic Reasoning and Tool Allocation}, where reasoning requirements evolve throughout execution as new observations are obtained from tools, memory, or the environment, requiring adaptive mechanisms that jointly allocate reasoning effort, tool usage, and token budgets according to task complexity while avoiding both over- and under-reasoning.

Future research should focus on adaptive reasoning mechanisms that dynamically determine when to reason, act, and stop by balancing reasoning cost, uncertainty, task complexity, and the expected benefit of additional computation.

\paragraph{A Candidate Adaptive-Reasoning Controller.}
One promising direction is the development of an adaptive-reasoning controller that dynamically allocates reasoning effort throughout an agent's execution. Rather than relying on fixed reasoning budgets, the controller would determine whether to continue reasoning, invoke external tools, or produce a final answer based on the current reasoning state, task requirements, and the expected benefit of additional computation. Such a controller could balance reasoning quality against computational cost, enabling agents to allocate reasoning only when it is likely to improve task performance.

\paragraph{A Measurable Evaluation Suite.}
More broadly, reasoning-efficient agentic AI will require new benchmarks, evaluation metrics, and control mechanisms that explicitly measure and optimize reasoning allocation rather than simply increasing reasoning effort. Progress toward adaptive reasoning therefore requires evaluation beyond task accuracy alone. Future work could consider an evaluation suite that measures how effectively reasoning effort is allocated, including reasoning efficiency (accuracy per reasoning token), stopping behavior, unnecessary tool usage, planning drift, and computational cost. Such metrics could enable systematic evaluation of reasoning allocation strategies and reveal trade-offs that are not captured by answer accuracy alone.

\section{Conclusion}

In this paper, we investigated reasoning efficiency in agentic AI and identified over-reasoning and under-reasoning as recurring failure modes of reasoning allocation. Our findings suggest that cases classified as over-reasoning are associated with higher latency, greater token consumption, more frequent tool usage, and increased token-limit exhaustion without proportional improvements in task performance, whereas cases classified as under-reasoning are consistently associated with incorrect or incomplete solutions under our evaluation rubric. Although MATH-500 and GAIA exhibit different dominant failure modes, both benchmarks indicate that the evaluated agent configuration does not always allocate reasoning effort and tool usage according to task requirements. These findings suggest that the goal of future agentic AI systems should not be to maximize reasoning effort, but to allocate reasoning adaptively according to evolving task demands. We hope this work motivates future research on adaptive reasoning mechanisms that dynamically determine when to reason, when to act, and when to stop.

\section{Limitations}

This work presents a preliminary empirical study of reasoning allocation in a LangGraph-based agentic AI framework using a fixed Qwen3.5-4B tool-routing model and three final-response models. Consequently, our findings should not be generalized to all agentic architectures, routing strategies, or foundation models. Although reasoning regimes are independently classified from answer correctness, our analysis does not explicitly distinguish tool-routing failures, retrieval failures, reasoning failures, response synthesis errors, or formatting errors, all of which may contribute to incorrect predictions. We evaluate inference cost using latency, token consumption, tool usage, and token-limit hits, but do not measure monetary cost, GPU utilization, energy consumption, or throughput. Furthermore, our experiments use a fixed maximum generation length of 4,096 tokens and do not include token-budget ablations, no-tool baselines, or standalone LLM baselines. Finally, we report average statistics only; future work should incorporate confidence intervals, distributional analyses, qualitative trace studies, and broader evaluations across additional agentic frameworks, models, and tasks.

\section*{Ethics and Reproducibility Statement}
This work investigates the reasoning challenges that emerge when LLMs are integrated into agentic AI systems. Our analysis identifies over-reasoning and under-reasoning as recurring failure modes that can increase computational cost or degrade task performance, motivating future research on more adaptive reasoning mechanisms. To support reproducibility, we will release the complete source code for our LangGraph-based framework, experimental configurations, prompts (including the GPT-4.1 judge prompt and reasoning-classification rubric), tool definitions, and per-sample evaluation outputs.

\section*{Acknowledgement}

This work is partly based upon the work supported by the National Center for Transportation Cybersecurity and Resiliency (TraCR) (a U.S. Department of Transportation National University Transportation Center) headquartered at Clemson University, Clemson, South Carolina, USA. Any opinions, findings, conclusions, and recommendations expressed in this material are those of the author(s) and do not necessarily reflect the views of TraCR, and the U.S. Government assumes no liability for the contents or use thereof.

This research is also supported by the National Artificial Intelligence Research Resource (NAIRR) Pilot and  AWS through the CloudBank project, which is supported by National Science Foundation grant\#1925001. Any opinions, findings, conclusions, and recommendations expressed in this material are those of the author(s) and do not necessarily reflect the views of  NAIRR, and the U.S. Government assumes no liability for the contents or use thereof.

\bibliographystyle{ACM-Reference-Format}
\bibliography{References}

@article{sui2025stop,
  title={Stop overthinking: A survey on efficient reasoning for large language models},
  author={Sui, Yang and Chuang, Yu-Neng and Wang, Guanchu and Zhang, Jiamu and Zhang, Tianyi and Yuan, Jiayi and Liu, Hongyi and Wen, Andrew and Zhong, Shaochen and Zou, Na and others},
  journal={arXiv preprint arXiv:2503.16419},
  year={2025}
}

@article{aggarwal2025optimalthinkingbench,
  title={Optimalthinkingbench: Evaluating over and underthinking in llms},
  author={Aggarwal, Pranjal and Kim, Seungone and Lanchantin, Jack and Welleck, Sean and Weston, Jason and Kulikov, Ilia and Saha, Swarnadeep},
  journal={arXiv preprint arXiv:2508.13141},
  year={2025}
}

@article{sapkota2025ai,
  title={Ai agents vs. agentic ai: A conceptual taxonomy, applications and challenges},
  author={Sapkota, Ranjan and Roumeliotis, Konstantinos I and Karkee, Manoj},
  journal={Information Fusion},
  pages={103599},
  year={2025},
  publisher={Elsevier}
}

@inproceedings{wu2025agentic,
  title={Agentic reasoning: A streamlined framework for enhancing llm reasoning with agentic tools},
  author={Wu, Junde and Zhu, Jiayuan and Liu, Yuyuan and Xu, Min and Jin, Yueming},
  booktitle={Proceedings of the 63rd Annual Meeting of the Association for Computational Linguistics (Volume 1: Long Papers)},
  pages={28489--28503},
  year={2025}
}

@inproceedings{kim2026cost,
  title={The cost of dynamic reasoning: Demystifying ai agents and test-time scaling from an ai infrastructure perspective},
  author={Kim, Jiin and Shin, Byeongjun and Chung, Jinha and Rhu, Minsoo},
  booktitle={2026 IEEE International Symposium on High Performance Computer Architecture (HPCA)},
  pages={1--16},
  year={2026},
  organization={IEEE}
}

@article{tran2026single,
  title={Single-agent llms outperform multi-agent systems on multi-hop reasoning under equal thinking token budgets},
  author={Tran, Dat and Kiela, Douwe},
  journal={arXiv preprint arXiv:2604.02460},
  year={2026}
}

@article{wang2025efficient,
  title={Efficient agents: Building effective agents while reducing cost},
  author={Wang, Ningning and Hu, Xavier and Liu, Pai and Zhu, He and Hou, Yue and Huang, Heyuan and Zhang, Shengyu and Yang, Jian and Liu, Jiaheng and Zhang, Ge and others},
  journal={arXiv preprint arXiv:2508.02694},
  year={2025}
}

@inproceedings{becker2026stay,
  title={Stay focused: Problem drift in multi-agent debate},
  author={Becker, Jonas and Kaesberg, Lars Benedikt and Stephan, Andreas and Wahle, Jan Philip and Ruas, Terry and Gipp, Bela},
  booktitle={Findings of the Association for Computational Linguistics: EACL 2026},
  pages={5068--5102},
  year={2026}
}

@article{hosseini2025role,
  title={The role of agentic ai in shaping a smart future: A systematic review},
  author={Hosseini, Soodeh and Seilani, Hossein},
  journal={Array},
  volume={26},
  pages={100399},
  year={2025},
  publisher={Elsevier}
}

@article{acharya2025agentic,
  title={Agentic AI: Autonomous intelligence for complex goals—A comprehensive survey},
  author={Acharya, Deepak Bhaskar and Kuppan, Karthigeyan and Divya, B},
  journal={IEEe Access},
  volume={13},
  pages={18912--18936},
  year={2025},
  publisher={IEEE}
}

@article{zhou2025exploring,
  title={Exploring the Necessity of Reasoning in LLM-based Agent Scenarios},
  author={Zhou, Xueyang and Tie, Guiyao and Zhang, Guowen and Wang, Weidong and Zuo, Zhigang and Wu, Di and Chu, Duanfeng and Zhou, Pan and Gong, Neil Zhenqiang and Sun, Lichao},
  journal={arXiv preprint arXiv:2503.11074},
  year={2025}
}

@article{su2025between,
  title={Between underthinking and overthinking: An empirical study of reasoning length and correctness in llms},
  author={Su, Jinyan and Healey, Jennifer and Nakov, Preslav and Cardie, Claire},
  journal={arXiv preprint arXiv:2505.00127},
  year={2025}
}

@article{zhou2026more,
  title={When More Thinking Hurts: Overthinking in LLM Test-Time Compute Scaling},
  author={Zhou, Shu and Ling, Rui and Chen, Junan and Wang, Xin and Fan, Tao and Wang, Hao},
  journal={arXiv preprint arXiv:2604.10739},
  year={2026}
}

@article{wei2022chain,
  title={Chain-of-thought prompting elicits reasoning in large language models},
  author={Wei, Jason and Wang, Xuezhi and Schuurmans, Dale and Bosma, Maarten and Xia, Fei and Chi, Ed and Le, Quoc V and Zhou, Denny and others},
  journal={Advances in neural information processing systems},
  volume={35},
  pages={24824--24837},
  year={2022}
}

@article{yao2022react,
  title={React: Synergizing reasoning and acting in language models},
  author={Yao, Shunyu and Zhao, Jeffrey and Yu, Dian and Du, Nan and Shafran, Izhak and Narasimhan, Karthik and Cao, Yuan},
  journal={arXiv preprint arXiv:2210.03629},
  year={2022}
}

@techreport{openai2024o1,
  title={Learning to Reason with LLMs},
  author={{OpenAI}},
  institution={OpenAI},
  year={2024},
  url={https://openai.com/index/learning-to-reason-with-llms/}
}

@article{amini2025distributed,
  title={Distributed llms and multimodal large language models: A survey on advances, challenges, and future directions},
  author={Amini, Hadi and Mia, Md Jueal and Saadati, Yasaman and Imteaj, Ahmed and Nabavirazavi, Seyedsina and Thakker, Urmish and Hossain, Md Zarif and Fime, Awal Ahmed and Iyengar, SS},
  journal={arXiv preprint arXiv:2503.16585},
  year={2025}
}

@inproceedings{mialon2024gaia,
  title={Gaia: a benchmark for general ai assistants},
  author={Mialon, Gr{\'e}goire and Fourrier, Cl{\'e}mentine and Wolf, Thomas and LeCun, Yann and Scialom, Thomas},
  booktitle={International Conference on Learning Representations},
  volume={2024},
  pages={9025--9049},
  year={2024}
}

@misc{math500,
  title        = {MATH-500},
  author       = {Hugging Face H4 Team},
  year         = {2024},
  howpublished = {\url{https://huggingface.co/datasets/HuggingFaceH4/MATH-500}},
  note         = {500-problem evaluation subset of the MATH benchmark}
}

@article{abdin2025phi,
  title={Phi-4-reasoning technical report},
  author={Abdin, Marah and Agarwal, Sahaj and Awadallah, Ahmed and Balachandran, Vidhisha and Behl, Harkirat and Chen, Lingjiao and de Rosa, Gustavo and Gunasekar, Suriya and Javaheripi, Mojan and Joshi, Neel and others},
  journal={arXiv preprint arXiv:2504.21318},
  year={2025}
}

\end{document}